\documentclass{article}
\usepackage[nonatbib]{nips_2017}
\usepackage{cite}
\usepackage{amsmath,amssymb,amsfonts}
\usepackage{mathrsfs}
\usepackage{algorithmic}
\usepackage{graphicx}
\usepackage{textcomp}
\usepackage{xcolor}
\usepackage{url} 

\def\BibTeX{{\rm B\kern-.05em{\sc i\kern-.025em b}\kern-.08em
    T\kern-.1667em\lower.7ex\hbox{E}\kern-.125emX}}
\begin{document}

\title{Differentiable lower bound for expected BLEU score}

\author{
  Vlad Zhukov\\
  Neural Networks and Deep Learning Lab\\
  Moscow Institute of Physics and Technology\\
  Russia \\
  \texttt{vladislav.a.zhukov@phystech.edu} \\
  \And
  Eugene Golikov\\
  Neural Networks and Deep Learning Lab\\
  Moscow Institute of Physics and Technology\\
  Russia \\
  \texttt{golikov.ea@mipt.ru} \\
  \And
  Maksim Kretov\\
  Neural Networks and Deep Learning Lab\\
  Moscow Institute of Physics and Technology\\
  Russia \\
  \texttt{kretov.mk@mipt.ru} \\
  }

\maketitle

\begin{abstract}
  Performance of the models in natural language processing (NLP) tasks is often measured with some non-differentiable metric, such as BLEU score. To use efficient gradient-based methods for optimization it is a common workaround to optimize some surrogate loss function. This approach is effective if optimization of such loss also results in improving target metric. Discrepancy between metric being optimized and target metric is referred to as loss-evaluation mismatch. In the present work we propose a new technique for calculation of differentiable lower bound (LB) of BLEU score. It does not involve computationally expensive sampling procedure such as the one required when using REINFORCE rule from reinforcement learning (RL) framework. 
\end{abstract}

\section{Introduction}
\label{intro}
One of the features of NLP tasks is complicated quality metrics. We often cannot design and calculate such metrics rigorously. For example, it is hard to quantify readability of text as perceived by human evaluator. So benchmark metrics are used: BLEU \cite{bleu}, ROUGE \cite{rouge}. Although these metrics do not address the problem mentioned above in full \cite{eval_empir}, they are interpretable and easy to compute.

A problem with these metrics is that they are usually non-differentiable hence gradient-based methods for efficient optimization cannot be applied directly. This complication can be addressed at least in two principal ways:

\begin{itemize}
  \item Design differentiable surrogate loss function which provides intuition and correlates with target metric. An example is optimization of cross-entropy loss with teacher forcing \cite{teacher_forcing}. Zero cross-entropy loss means target and generated texts are exactly the same.
  \item Appeal to RL-like techniques (REINFORCE rule \cite{reinforce} and its generalizations \cite{scg}) in order to optimize expected value of target metric \cite{mrt,abs_sumariz}. This approach is typically data-inefficient, and a number of practical tweaks are required to ensure reasonably fast convergence \cite{var_red}.
\end{itemize}

In the present article both approaches are combined. We design differentiable surrogate loss function which serves as a lower bound for the expected value of BLEU score. Thus we eliminate sampling procedure from optimization technique (as in REINFORCE) at the cost of optimizing LB instead of an exact expected value of target metric. The computational graph becomes fully differentiable. We illustrate benefits of the proposed approach on simple translation tasks.

\section{BLEU score}
\label{bleu}
\subsection{Definition}

Let us remind how BLEU score for the pair of candidate text $C$ and reference text $R$ is calculated \cite{bleu}. Two key terms are needed: n-gram precisions $p_n$ and brevity penalty BP. Precisions are calculated as follows:
\begin{equation}
p_n = \frac{O_n}{\sum\limits_{\textnormal{n-gram} \in C} {\textnormal{Count$_C$(n-gram)}}}
\label{p_n}
\end{equation}
For simplicity, we only consider single candidate text $C$ instead of a set of few possible candidate texts $\{C\}$ as in the original definition. Thus denominator is a number of all possible n-grams in $C$. For unigrams that would be just a length of $C$. The numerator contains so-called overlaps $O_n$ between candidate and reference texts:
\begin{equation}
\sum\limits_{\textnormal{n-gram} \in C} \textnormal{min(Count$_C$(n-gram), Count$_R$(n-gram))}
\label{Oprime}
\end{equation}

Summation is over all n-grams from candidate text $C$. Calculation of brevity penalty:
\begin{equation}
\textrm{BP} =
\begin{cases}
1 & \text{if}\ c>r \\
e^{1 - r/c} & c \le r
\end{cases}
\end{equation}

Here $r$ is the length of $R$, and $c$ is the length of $C$. BLEU score is calculated as geometric mean of n-gram precisions with weights $w_n$:
\begin{equation}
\textrm{BLEU}=\textrm{BP} \cdot \exp \left(\sum\limits_{n=1}^N{w_n \log p_n}\right)
\end{equation}

Typical parameter set is: $w_n = 1/N, N = 4$.

\subsection{Matrix form}
To facilitate further derivation we rewrite the above-mentioned formulae for BLEU score in matrix form. Without loss of generality we consider pair of a candidate text $C$ and a corresponding reference text $R$ with vocabulary of size $v$. We denote by $x$ the matrix with shape $\lbrack \textnormal{len$_x$}\times v \rbrack$ which contains one-hot encoded vectors for each word in $C$. Similarly, we denote by $y$ the matrix with shape $ \lbrack \textnormal{len$_y$} \times v \rbrack$ which contains one-hot encoded vectors for each word in $R$.

Now we compute intermediate matrices $S^{ 1}=xx^{T}$ and $P^{ 1}=yx^{T}$
with shapes $\lbrack \textnormal{len$_x$} \times \textnormal{len$_x$} \rbrack$
and $ \lbrack \textnormal{len$_y$} \times \textnormal{len$_x$} \rbrack,$ correspondingly:
$$
S^{ 1}_{i, j} = x_i x_j
\qquad
P^{ 1}_{i, j} = y_i x_j
$$

Upper index denotes $n$ in n-grams. In the similar manner matrices for all n-grams with $n>1$ are defined:
\begin{equation}
S^{ n}_{i, j} = \prod\limits_{k=0}^{n-1}x_{i + k} x_{j + k}
\qquad
P^{ n}_{i, j} = \prod\limits_{k=0}^{n-1}y_{i + k} x_{j + k}
\end{equation}

Corresponding shapes are $\lbrack \textnormal{len$_x$} - n + 1 \times \textnormal{len$_x$} - n + 1 \rbrack$ and $\lbrack \textnormal{len$_y$} - n + 1 \times \textnormal{len$_x$} - n + 1 \rbrack$. Elements of these matrices have the following form:
\begin{equation}
 S^{ n}_{i, j}=
  \begin{cases}
    1 & \text{if}\ C_i^n = C_j^n \\
    0 & \textrm{otherwise}
  \end{cases}
\end{equation}
\begin{equation}
P^{ n}_{i, j}=
 \begin{cases}
   1 & \text{if}\ R_i^n = C_j^n \\
   0 & \textrm{otherwise}
 \end{cases}
\end{equation}

Here $C_i^n(R_i^n)$ is n-gram on the position $i$ in $C(R)$ text: to get this n-gram we need to take $n$ words starting from $i$-th position in the text. 

For every n-gram taken from specific position in $C$ we count same n-grams in whole $C$ or $R$ text:
\begin{equation}
v^{x, n}_{i} =  \sum\limits_{j=0}^{\textnormal{len$_x$} - n} S^{n}_{j, i}
\qquad
v^{y, n}_{i} = \sum\limits_{j=0}^{\textnormal{len$_y$} - n} P^{n}_{j, i}
\end{equation}
These are just counters obtained by summation over columns in original matrices $S^n$ and $P^n$. Basically, $i$-th element of vector $v^{x, n}$ shows how many times n-gram on $i$-th position in $C$ occur in whole $C$ text. Illustrative example is given in Appendix A.1.

Next we combine these vectors to calculate overlap $O_n$:
\begin{equation}
  O_n = \sum\limits_{i = 0}^{\textnormal{len}_x-n} O_n^i = 
  \sum\limits_{i = 0}^{\textnormal{len}_x-n} \frac{\min(v^{x, n}_i, v^{y, n}_i)}{v^{x, n}_i} =
\label{o_n_2}
\end{equation}
$$ \sum\limits_{i = 0}^{\textnormal{len}_x-n} \min(1, \frac{v^{y, n}_i}{v^{x, n}_i}) = $$
\begin{equation}
  \sum\limits_{i = 0}^{\textnormal{len}_x-n} \min \Big(1,
  \frac{\sum_{j=0}^{\textnormal{len$_y$}-n} \prod_{k=0}^{n-1}y_{i + k} x_{j + k}}{\sum_{j=0}^{\textnormal{len$_x$}-n} \prod_{k=0}^{n-1}x_{i + k} x_{j + k}}\Big) 
\end{equation}

See Appendix A for the technical proof that Eqs. (\ref{Oprime}) and (\ref{o_n_2}) are equivalent. Calculation of the corresponding precisions is straightforward:
\begin{equation}
p_n = \frac{O_n}{\textnormal{len}_x-n+1}
\label{p_n_2}
\end{equation}

Thus we derived formula for calculation of BLEU score that only uses matrices $x$ and $y$ as an input and all operations are differentiable in contrast with Eq. (\ref{Oprime}) where we had non-differentiable Count operation.

It is to be noted, that $x$ and $y$ are matrices of the special form comprised of stacked 1-hot vectors. These vectors are rows of the matrices and represent words sampled by translation model from output probability distributions. 

Recall that our objective is to maximize the expected BLEU score:
\begin{equation}
\mathbb{E}_{\mathrm{data}} \mathbb{E}_{\mathrm{arch\_stoch}} \mathbb{E}_{\mathrm{output}} \textnormal{BLEU}(x, y) \to \max,
\label{objective}
\end{equation}
Here the first expectation is taken over training dataset. Second one is 'architecture stochasticity' and corresponds to the inner stochasticity of the model. The most common example of such stochasticity is feeding decoder at time step $t$ with tokens sampled from distributions emitted at step $t-1$.

Third expectation occurred due to the fact that decoder in a typical sequence-to-sequence model \cite{seq2seq} emits the distributions over successive tokens, not the tokens themselves. BLEU score is calculated from samples and we want to optimize its expected value given emitted probability distributions on all time steps (matrix $p_x$). 

Please note, that we disentangled here sampling from word distributions to produce tokens fed to next steps of decoder and sampling which is used for calculation of BLEU score.

Given these three expectations, optimization of expected BLEU score is a challenging problem which is a focus of the next section. In the next section we first discuss the most common approach of direct optimization with RL techniques and then move to analytical derivation of LB for expected BLEU score.

\section{Direct optimization of BLEU score}
\subsection{REINFORCE rule}

BLEU score is calculated given $C$ and $R$. They consist of the discrete tokens and BLEU score is not differentiable respect to inputs: we re-formulated non-differentiable Count operation but still matrices $x$ and $y$ cannot be varied smoothly. Common approach in these cases is to consider probability distributions from which $x$ and $y$ were sampled and use techniques from RL, specifically REINFORCE rule \cite{reinforce} for optimization of the expected value of metric of interest. General formula for REINFORCE rule:
$$
J(\theta) = \mathbb{E}_{\tau\sim \pi_{\theta}} \lbrack R(\tau) \rbrack
$$
$$
\Rightarrow \nabla_\theta J(\theta) = \mathbb{E}_{\tau} \Big\lbrack \sum_t R_t \nabla_\theta\log \pi_{\theta}(a_t|s_t) \Big\rbrack
$$

Here $R(\tau)$ is a total reward that agent accumulates from all time steps during trajectory $\tau$: $R(\tau)=\sum_t R_t$. It is our target metric that we want to maximize during training. In the context of NLP tasks trajectory can be seen as sequence of emitted words and hidden states generated by decoder: $(w_1, h_1,..w_N, h_N)$. Reward $R(\tau)$ corresponds to BLEU score and is calculated on the basis of emitted words. Stochastic policy $\pi_\theta(a_t|s_t)$ forms a probability distribution over the words at time step $t$. 

Problem with this formula is that random variable in integrand usually shows high variance \cite{var_red}. Sample-efficient estimation of this expectation is challenging and requires a number of tricks \cite{var_red}, such as control variates, see for example \cite{rebar}.

While for typical RL tasks this integral is indeed very complex and requires numerical estimation, we reckon that in some specific tasks, such as calculation of BLEU score, analytical formula for LB on this integral can be derived in the closed form. 

More precisely, our idea is to provide an analytical lower bound for the last expectation in Eq.~\ref{objective}. Using this lower bound allows us to not to estimate this expectation via Monte-Carlo procedure, as we would have to, if we choose to use REINFORCE rule. Hence we expect the variance of the gradient of the objective to be lower than the corresponding variance of REINFORCE. The gradient of the objective with remaining two expectations can be estimated using stochastic computation graph formalism~\cite{scg}.

\subsection{LB for expected BLEU score}

In order to proceed we have to make the following simplifying assumption: all the words in $R$ are unique and encoded as one-hot vectors.
We denote by $p_y$ matrix of stacked one-hot vectors that represent reference text $R$ (degenerate distribution). 

High-level plan of deriving the LB:
\begin{enumerate}
  \item Write down explicit formula for mathematical expectation of BLEU score using distributions $p_x$ and degenerate distribution $p_y$.
  \item Introduce assumptions that allow deriving analytical formula for LB on the integral.
  \item Show that in case of degenerate distribution $p_x$ derived LB coincides with exact BLEU score.
  \item Consider values of $p_x$ as parameters of loss function (negative LB in our case). Derivatives of loss function w.r.t. parameters can be calculated with automatic differentiation software \cite{pytorch}. Demonstrate efficiency on toy examples.
\end{enumerate}

Let us now elaborate the plan. Hereinafter we denote expected score as BLEU and score for specific input texts $x$ (sampled from $p_x$) and $y$ (deterministic sample from $p_y$ due to degeneracy of distribution) as $\textrm{BLEU}(x,y)$. First write down formula for BLEU:
\begin{equation}
\begin{split}
\textrm{BLEU} = \mathbb{E}_{x\sim p_x,y \sim p_y} \lbrack \textrm{BLEU}(x, y) \rbrack =\\
\mathbb{E}_{x\sim p_x,y \sim p_y} \Big \lbrack  \exp \Big(\sum_{n=1}^N \omega_n \log p_n \Big) \Big \rbrack
\label{gen_rbleu}
\end{split}
\end{equation}

Distribution over $y$ is omitted for brevity hereinafter. BLEU score is expectation of product of individual factors corresponding to n-grams. It can be bounded by the product of expectations:
\begin{equation}
\begin{split}
\textrm{BLEU} = \mathbb{E}_{x \sim p_x}  \Bigg \lbrack \exp\left(\sum\limits_{n=1}^N{w_n \log p_n} \right) \Bigg \rbrack \\
\geq  \prod_{n=1}^{N}  \big(\mathbb{E}_{x \sim p_x} \lbrack p_n \rbrack \big)^{\omega_n} - 1
\label{cov}
\end{split}
\end{equation}

Last estimate is trivial given that precisions $p_n$ are non-negative. Let us now consider single factor in this product. According to Eq.(\ref{p_n_2}), $p_n$ is proportional to overlaps $O_n$. So, for unigrams (BLEU1):
\begin{equation}
\begin{split}
\mathbb{E}_{x\sim p_x} \big\lbrack \sum_{i=0}^{len_x-1}{O_1^i} \big\rbrack =
\mathbb{E}_{x\sim p_x} \Big \lbrack \sum_{i=0}^{len_x-1} \frac{\min(v^{y, 1}, v^{x,1})}{v^{x, 1}} \Big \rbrack
\label{bleu_1}
\end{split}
\end{equation}

Using Jensen's inequality and definitions of $v^{x,n},v^{y,n}$ we obtain LB for $i$-th component of n-gram overlap (full derivation in Appendix B):
\begin{equation}
  \mathbb{E}_{x\sim p_x} \big\lbrack O_1^i \big\rbrack \geq \sum_n p_x^{in} \cdot \min \big(1, \frac{\sum_j y_{jn}}{1+\sum_{k\neq i} p_x^{kn}}\big)
\end{equation}

Thus, LB for components of overlap $O_1$ is easily calculated using $p_x$ and reference text. LB for n-grams with $n>1$ is obtained in the similar manner (Appendix C):
\begin{equation}
\begin{split}
  \label{bleu_n}
  \mathbb{E}_{x\sim p_x} \big[O_n^i \big] \geq   \sum_{m_0} p_x^{i,m_0} .. \sum_{m_{n-1}} p_x^{i+n-1,m_n} \cdot \\
 \cdot \min \big(1, \frac{\sum_j \prod_{k=0}^{n-1} y_{j + k, m_k}}{1 + \sum_{l \ne i} \prod_{k=0}^{n-1} p_x^{l+k,m_k}} \big)
\end{split}
\end{equation}

Derived LB for overlap of $n$-th order is tight in the sense that if $p_x$ and $p_y$ are both deterministic distributions (probabilities of words are either 1 or 0) then LB coincides with exact value. So if we want to optimize just one type of n-grams (say, 4-grams only) and not the aggregated BLEU score, we can use Eq. \ref{bleu_n}. 

If we optimize the usual BLEU score with all n-grams up to order $n$ simultaneously, we use product of such lower bounds, but this product does not coincide with the exact BLEU score even for degenerate distribution (bound in Eq.\ref{cov} is too coarse).

\section{Results and discussion}
In this section we first demonstrate high correlation between LB and actual expected BLEU score on the toy examples. Next we demonstrate benefits of the proposed approach on the simple translation task.
\begin{figure}[h]
  \centering
  \includegraphics[scale=0.16]{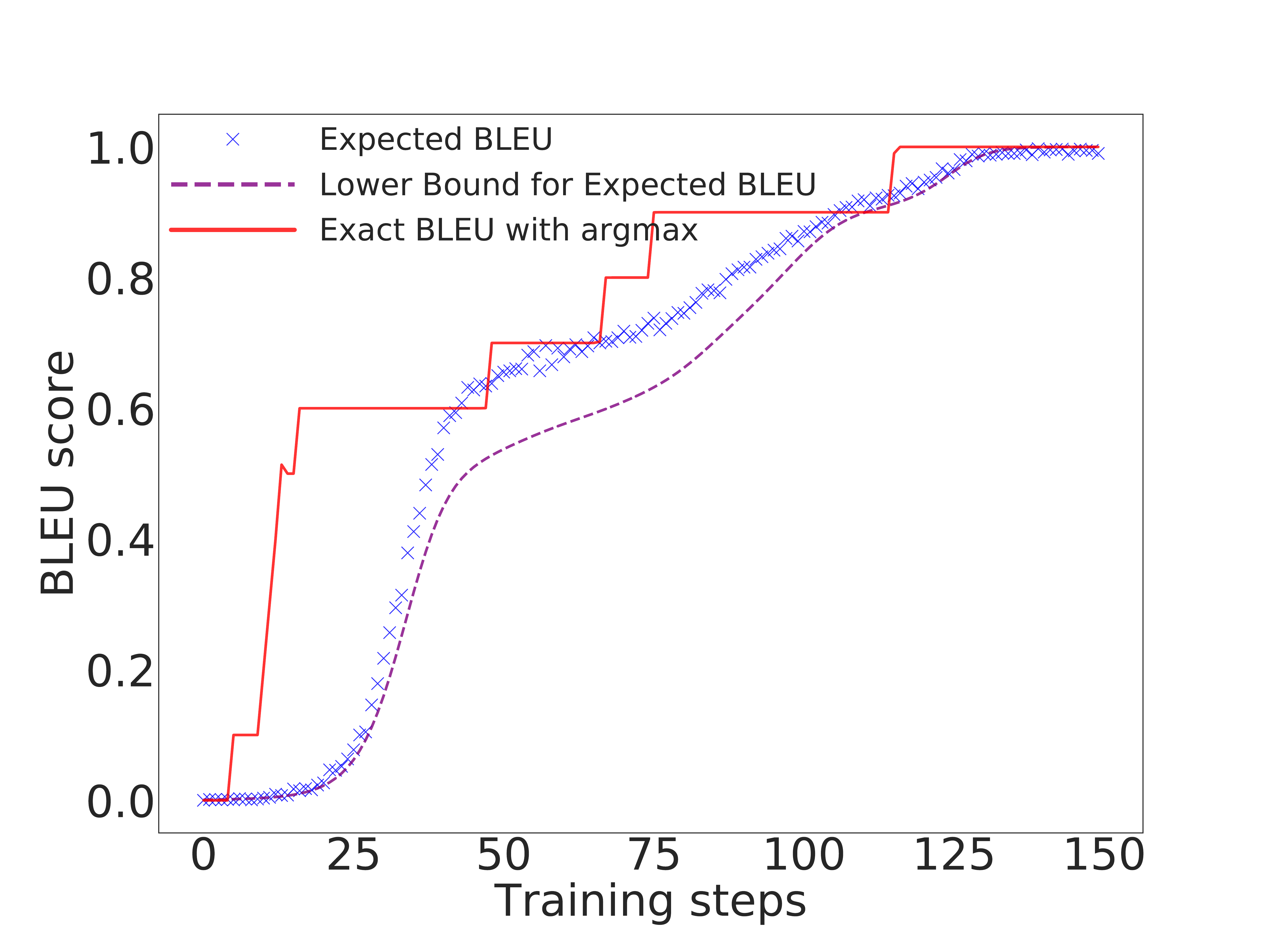}
  \caption{Example of learning curve for unigram BLEU1 score. Red line is exact BLEU score obtained by applying $argmax$ to $p_x$ row-wise and calculating BLUE score. Purple dotted line is lower bound for expected BLEU score. Crosses denote expected BLEU score calculated by averaging samples from distribution $p_x$.}
  \label{fig:bleu_toy}
\end{figure}

\begin{figure}[h]
  \centering
  \includegraphics[scale=0.16]{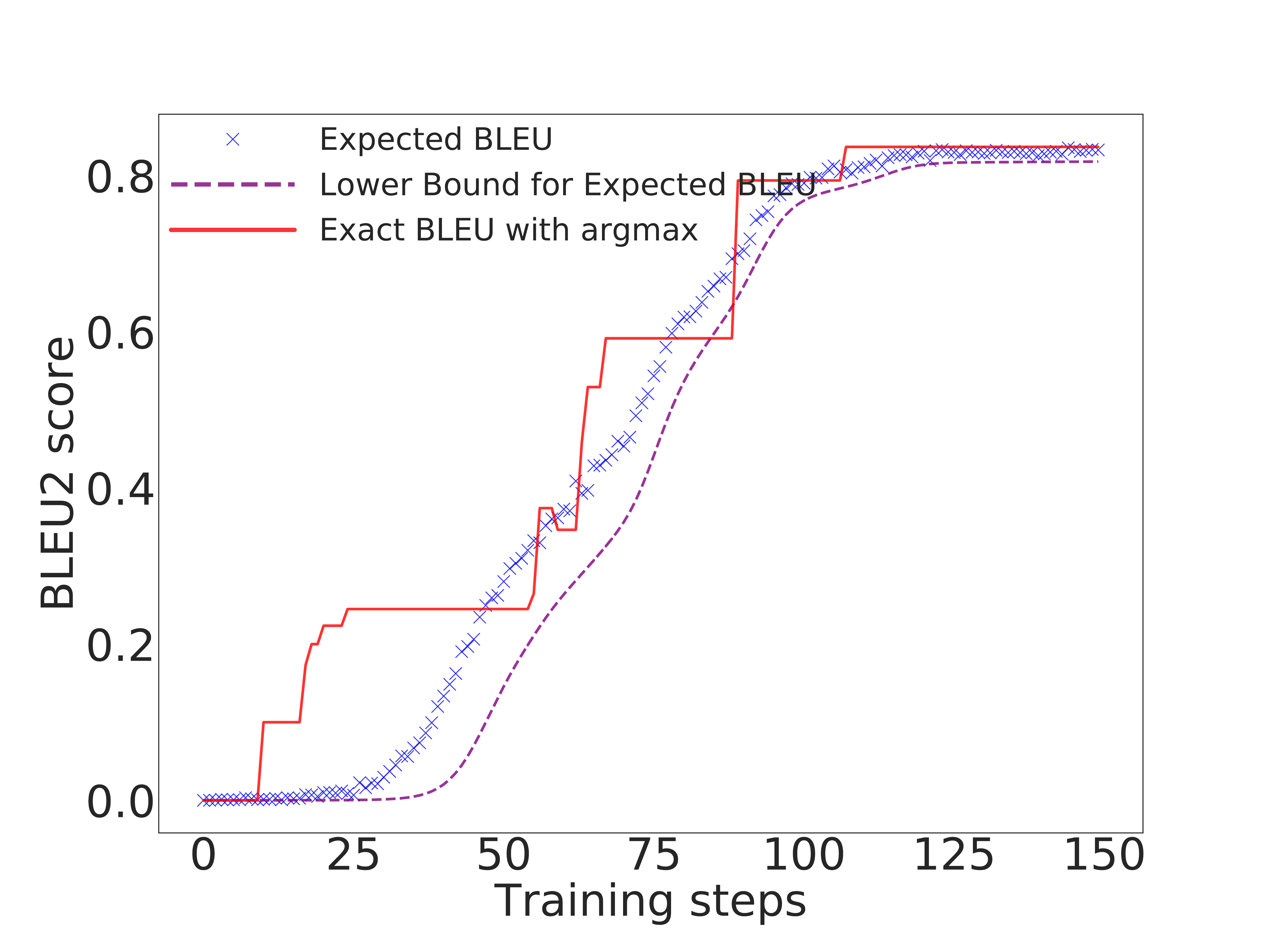}
  \caption{Example of learning curve for bigram BLEU2 score. Crosses show sum $\textrm{LB}+1$.}
  \label{fig:bleu2_toy}
\end{figure}
\subsection{Toy task}
To test our approach we designed the following toy task: generate matrix of candidate word distributions $p_x$ and matrix of one-hot vectors of reference text $r$. Both texts are of the same length 10. Vocabulary size is set to 10000. Training examples are generated randomly:
\begin{itemize}
  \item Elements of matrix $p_x$ are generated by sampling logits from $\mathcal{N} (0,1)$ and then $softmax$ function is applied row-wise to obtain probability distributions over words.
  \item Indices of reference text $R$ are generated by uniform sampling with replacement from vocabulary. 
\end{itemize}

In such setting, our target is to optimize parameters (elements of matrix $p_x$) to maximize expected BLEU score between two corpora of text: one is sampled from $p_x$ and another is $r$. For optimization we used lower bound of expected BLEU score.

Learning curve is presented at Figure \ref{fig:bleu_toy}. Chart indicates highly correlated behavior of proposed LB and expected BLEU score.

\subsection{Translation task}
We tested proposed approach on two tasks with datasets of different size:

\begin{enumerate}
  \item IWSLT'14 dataset~\cite{iwslt14} of German-English sentence pairs. Same preprocessing of sentences as in~\cite{bso} was performed.
  \item WMT'14~\cite{wmt14} dataset of German-English sentence pairs. Preprocessed data from~\cite{attn} were used. This dataset is much bigger than IWSLT'14 (4.5M vs 0.2M training sentence pairs).
\end{enumerate} 

The following sections describe training protocol. Goal of these experiments was to demonstrate benefits of addressing loss-evaluation mismatch even for the simple translation models. Architectures were designed in such a way to keep reasonable balance between quality and fast experimentation (few hours of training on high-end GPU).

\begin{figure}[h]
  \centering
  \includegraphics[scale=0.17]{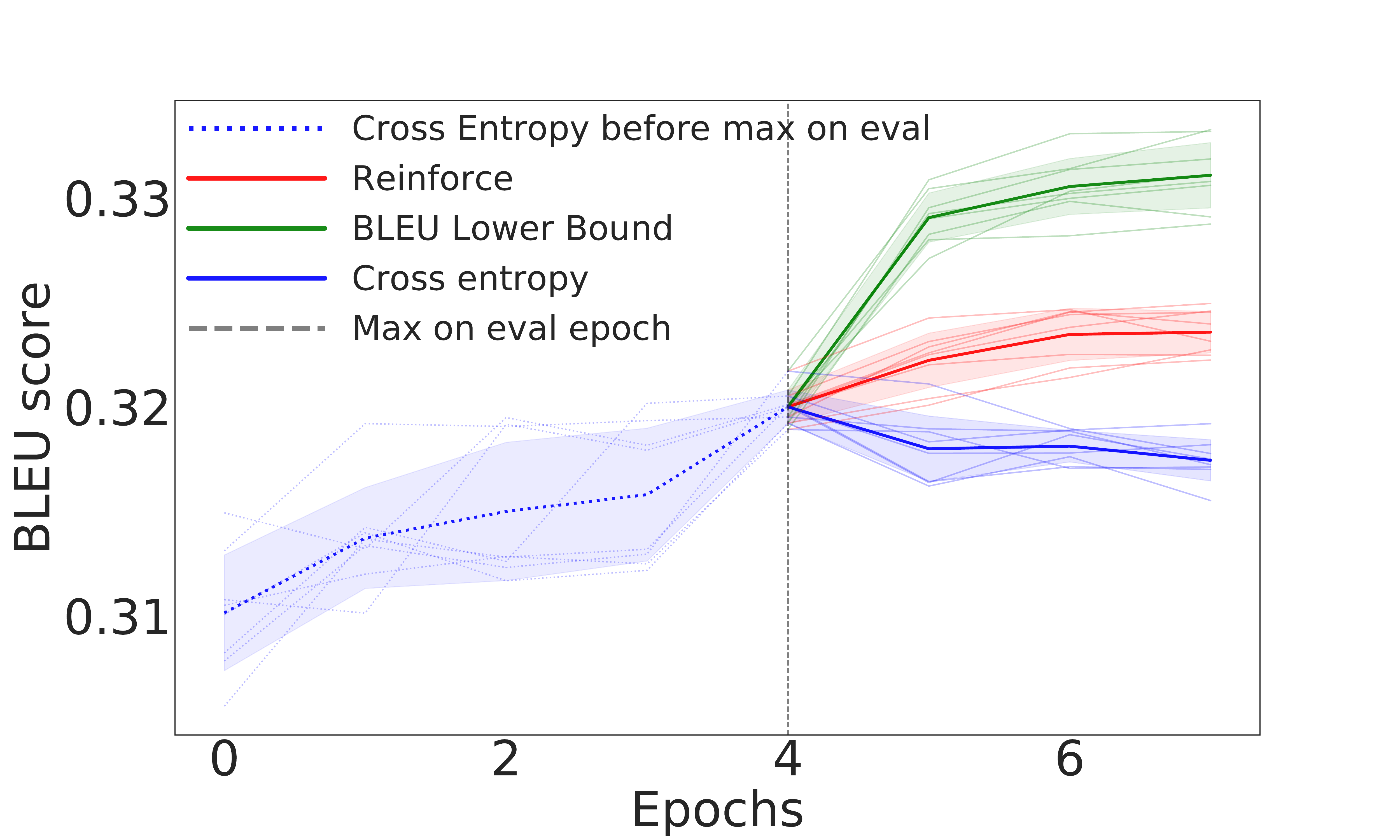}
  \caption{Training process for different optimization techniques, dataset IWSLT'14 DE-EN \cite{iwslt14}. At first model was optimized with cross-entropy loss. Starting from best weights (performance measured on validation set) model was further optimized with cross-entropy, LB and using REINFORCE rule. BLEU score is for evaluation dataset.}
  \label{fig:results}
\end{figure}

\begin{table}[t!]
\begin{center}
\begin{tabular}{|l|ccc|}
\hline \bf & \bf CE & \bf RF & \bf LB \\ \hline
1&$26.60\pm 0.14$&$27.16\pm 0.18$&$\textbf{27.83}\pm\textbf{0.17}$\\
5&$27.56\pm 0.17$&$27.85\pm 0.11$&$\textbf{28.10}\pm\textbf{0.17}$\\
\hline
\end{tabular}
\end{center}
\caption{\label{results_translation} Results for dataset IWSLT'14 DE-EN. Three optimization techniques: cross-entropy loss (CE), REINFORCE rule (RF), lower bound (LB) optimization (our approach). Beam sizes of 1 and 5. BLEU score is for test dataset.}
\end{table}

\begin{table}[t!]
\begin{center}
\begin{tabular}{|l|cc|}
\hline \bf & \bf CE  & \bf LB \\ \hline
1&$17.12\pm 0.2$&$\textbf{18.65}\pm \textbf{0.13}$\\
5&$\textbf{19.23}\pm \textbf{0.19}$&$19.03\pm 0.15$\\
\hline
\end{tabular}
\end{center}
\caption{\label{results_translation_wmt} Results for dataset WMT'14 DE-EN. Optimization with CE and LB. Beam sizes of 1 and 5. BLEU score is for test dataset.}
\end{table}

\subsubsection{Our model}
Standard sequence-to-sequence framework~\cite{seq2seq} was used. The encoder is a one-layer bi-directional LSTM \cite{lstm} with 256 hidden units for IWSLT'14 dataset and two-layer bi-directional LSTM with 512 hidden units for WMT'14. The decoder is one-layer LSTM with 256 hidden units for IWSLT'14 and two-layer bi-directional LSTM with 512 hidden units for WMT'14. In the decoder the multiplicative attention mechanism~\cite{attn} was implemented for calculation of context vector (weighted hidden states of encoder). Input fed into the decoder at step $t$ is a concatenation of the word embedding at step $t-1$ and a non-linear transformation of context vector and hidden state of the decoder at the previous step\footnote{Implementation is available at \url{github.com}}. During evaluation phase greedy decoding and beam search with beam size of 5 were tested.

\subsubsection{Training details}
We performed 8 runs in order to compute standard deviation of target metric. For optimization we used Adam \cite{adam} with the following starting learning rates: $\alpha=10^{-4}$ for cross-entropy loss and LB, $\alpha=10^{-5}$ for REINFORCE rule.

\subsubsection{Optimization protocol}
Training translation models consisted of four steps:
\begin{enumerate}
  \item Optimization of model with cross-entropy loss for 20 epochs.
  \item Selecting the best model weights using results on validation set.
  \item Further optimization of model for 5 epochs with one of the three approaches: cross-entropy loss, REINFORCE rule, LB.
  \item Evaluation on test set.
\end{enumerate}
We also found that additive smoothing of precisions increase stability of training on machine translation task:
$$
\textrm{LB}\Bigg \lbrack \frac{O_n}{\textnormal{len}_x - n + 1} \Bigg \rbrack \rightarrow\frac{ \textrm{LB} \lbrack O_n \rbrack + 1}{\textnormal{len}_x - n + 2}
$$
Results on test set are given in Table~\ref{results_translation} in the form "mean $\pm$ std". Optimization with LB loss resulted in higher BLEU scores on the test set. 

\section{Conclusions and future work}
We proposed a computationally efficient and differentiable lower bound for expected BLEU score. It is empirically shown on translation task that optimizing such objective results in improving target BLEU in comparison with alternatives (cross-entropy optimization, direct BLEU optimization with REINFORCE rule).

Proposed surrogate loss allows to eliminate data-inefficient sampling procedure from pipeline of direct optimization of target metric as in RL setting at the cost of optimizing lower bound instead of exact metric. By using proposed LB we also avoid selection of variance reduction techniques for REINFORCE rule, which may be a cumbersome exercise.

In order to use proposed LB only minor change to typical training protocol with cross-entropy loss and teacher forcing is needed: on later training epochs replace cross-entropy loss with LB loss. The implementation is available.

Main direction of future work is to combine proposed differentiable approximation for expected BLEU score with continuous relaxation of beam search optimization~\cite{cont_beam}. This direction is suggested by decreasing margin between proposed approach and cross-entropy training when using beam search with larger beam size for decoding (see Table~\ref{results_translation} and Table~\ref{results_translation_wmt}). Thus loss-evaluation mismatch will be eliminated and pipeline for training translation models become fully differentiable.

\bibliography{main}
\bibliographystyle{unsrt}

\appendix
\section{Proof of equivalence of formulas for overlaps $O_n$}
\label{app:OO}
Let $W$ denote set of unique n-grams in $C$, $I_{\omega}$ denote set of start indices for n-gram $\omega$ in $C$ (see example below). Then for overlap:
$$O_n = \sum\limits_{\omega \in W} \sum\limits_{i \in I_{\omega}}
\frac{\min(v_i^{y, n}, v_i^{x, n})}{v_i^{x, n}}$$
Let us consider n-gram $\omega$ from $C$. Suppose it occurs few times in $C$. If $\omega$ starts at $i$-th position in $C$ then $i$-th component of vector $v^{x, n}$ corresponds to number of occurrences of n-gram $\omega$ in $C$. Then by definition: 
$$
\begin{array}{ll}
\forall i, j \in I_{\omega}\\
v_i^{x, n} = v_j^{x, n} = |I_{\omega}| = Count_C(\omega)\\
v_i^{y, n} = v_j^{y, n} = Count_R(\omega)
\end{array}
$$
So we rewrite expression for overlap $O_n$:
$$O_n =\sum\limits_{\omega \in W} \sum_{i \in I_{\omega}} \frac{min(Count_R(\omega), Count_C(\omega))}{Count_C(\omega)} =$$ 
$$\sum\limits_{\omega \in W} min(Count_R(\omega), Count_C(\omega))$$
Last transition occurred because we have exactly $Count_C(\omega)$ identical terms in the summation.
\subsection{Example}
\label{app:O1}
Consider hypothesis
$$C = \textnormal{'Mike took an apple and the apple was delicious'}$$
Suppose we have unigram case $n=1$. Consider word $\omega=\textnormal{apple}$, then $I_w = \{3, 6\}$ is set of start indices of word 'apple' in given candidate text $C$. 

Number of occurrences of $\omega$ in $C$ is equal to 
$$
|I_w| = Count_C(w)=$$  $$=v_3^{x, 1} = v_6^{x, 1}=2
$$ 
by definition of vector $v^{x,1}$.

\section{Derivation of LB for unigram BLEU score}
\label{app:unigramLB}
Consider $i$-th component of overlap for unigrams (see Eq.~\ref{o_n_2} for definition of $O_n^i$):
\begin{equation}
\mathbb{E}_{x\sim p_x,y \sim p_y} O_1^i = \mathbb{E}_{x\sim p_x,y \sim p_y} \Big \lbrack \min \big(1, \frac{\sum_j x_i y_j}{\sum_k x_i x_k}\big) \Big \rbrack
\label{bleu_i_1}
\end{equation}

Break down expectation into two integrals, one is over random variable $x_i$ and the rest is over all other. Distribution over $y$ is omitted hereinafter:
$$
= \mathbb{E}_{x_{k,k\neq i}} \Big \lbrack \sum_n p_x^{in} \min \big(1, \frac{\sum_j y_{jn}}{\sum_k x_i x_k}\big) \Big \rbrack 
$$
\begin{equation}
= \sum_n p_x^{in} \cdot \mathbb{E}_{x_{k,k\neq i}} \Big \lbrack \min \big(1, \frac{\sum_j y_{jn}}{1+\sum_{k\neq i} x_i x_k}\big) \Big \rbrack
\label{bleu_i_2}
\end{equation}

Expression in square brackets can be considered as a function of variable $x_k$ with domain $[0,1]^n$. This function is convex (see Appendix \ref{app:convex}), so we can apply Jensen's inequality:
$$
= \sum_n p_x^{in} \cdot \mathbb{E}_{x_{k,k\neq i}} \Big \lbrack \min \big(1, \frac{\sum_j y_{jn}}{1+\sum_{k\neq i} x_i x_k}\big) \Big \rbrack
$$
\begin{equation}
\geq \sum_n p_x^{in} \cdot \min \big(1, \frac{\sum_j y_{jn}}{1+\sum_{k\neq i} p_x^{kn}}\big)
\label{bleu_i_3}
\end{equation}
Thus we derived LB for overlap $O_n=\sum_i O_n^i$.

\section{Generalization of LB for n-grams}
\label{app:n-grams}

We apply same logic as in Appendix~\ref{app:unigramLB} to derive LB for $O_n$. First write down expectation over random variables $\{x_{i + p}: p \in \{0..n-1\}\}$ for $i$-th component of overlap $O_n$:
$$
\mathscr{L}_i = {x_{k,k \not\in \{i + p: p \in \{0..n-1\} \} }}
$$
$$
\mathbb{E}_{x\sim p_x} \big[ O_n^i \big] =$$
\begin{equation}
\begin{split}
\mathbb{E}_{\mathscr{L}_i}
 \Big \lbrack
\sum_{m_0} p_x^{i,m_0} ... \sum_{m_{n-1}} p_x^{i+n-1,m_{n-1}}  O_n^i 
 \Big \rbrack
\label{bleu_i_2}
\end{split}
\end{equation}

Let $s$ be an index and $s \not\in \{i + p: p \in \{0..n-1\} \}$. 

Consider two sets: 
$$A_{s, i} = \{l: l \ne i, s \in [l..l+n-1]\}$$ 
$$B_{s, i}=\{l: l \ne i, s \not\in [l..l+n-1] \}$$
And rewrite expression for overlap term:
$$O_n^i = $$
$$
\min(1, \frac{\sum_j \prod\limits_{k=0}^{n-1}y_{k + k, m_k}} 
{1 + \sum\limits_{A_{s, i}} \prod\limits_{k=0}^{n-1} x_{i + k}x_{l + k}
+ \sum\limits_{B_{s, i}} \prod\limits_{k=0}^{n-1} x_{i + k}x_{l + k}})
$$
By changing order of integration and writing down separately expectation over $p_x^s$ we obtain:
\begin{equation}
\begin{split}
\sum_{m_0} p_x^{i,m_0} ... \sum_{m_{n-1}} p_x^{i+n-1,m_n} \\
\mathbb{E}_{x_{k,k \not\in \{i + p: p \in \{0..n-1\} \}, k \ne s} } \mathbb{E}_{p_x^s} O_n^i 
\label{bleu_i_3}
\end{split}
\end{equation}

Mathematical expectation over random variable $x^s$ has the following form:
$$
\min(1, \frac{A}{1 + B + \vec{c} x_s})
$$
Here $A, B$ are some constants and $\vec{c}$ is constant vector:
$$
\frac{A}{1 + B + \vec{c} x_s}
$$
Since this is convex function (see Appendix \ref{app:convex}), we can apply Jensen's inequality sequentially for all $s \not\in \{i + p: p \in \{0..n-1\} \}$. Let us define:
$$
O_i^{'n}=\min \big(1, \frac{\sum_j \prod_{k=0}^{n-1} y_{j + k, m_k}}{1 + \sum_{l \ne i} \prod_{k=0}^{n-1} p_x^{l+k,m_k}} \big)
$$
These new overlaps are functions only of probability distribution $p_x$ and matrix of reference text $y$. Thus we derived final inequality:
$$
\mathbb{E}_{x\sim p_x} \big[O_n^i \big]  \geq \sum_{m_0} p_x^{i,m_0} .. \sum_{m_{n-1}} p_x^{i+n-1,m_n} O_i^{'n}
\label{bleu_i_4}
$$
LB is a function of parameters of probability distributions and reference text. 

\section{Proof of convexity}
\label{app:convex}
Define $a$ as fixed vector, and $A, B$ are some positive constants:
\begin{equation}
\frac{A}{B + a((1 - \alpha)x_s + \alpha y_s)}
\le \frac{(1 - \alpha)A}{B + a x_s} + \frac{\alpha A}{B + a y_s}
\end{equation}
Without loss of generality:
\begin{equation}
\frac{1}{1 + a((1 - \alpha)x_s + \alpha y_s)}
\le \frac{1 - \alpha}{1 + a x_s} + \frac{\alpha}{1 + a y_s}
\end{equation}

Let us denote $S = a x_s$, $T = a y_s$:
$$\frac{1}{1 + (1 - \alpha)S + \alpha T} \le \frac{1 - \alpha}{1 + S} + \frac{\alpha}{1 + T}$$
$$\frac{1}{1 + (1 - \alpha)S + \alpha T} \le
\frac{(1 - \alpha)(1 + T) + \alpha (1 + S)}{(1 + T)(1 + S)}$$
$$\frac{1}{1 +  (1 - \alpha)S + \alpha T} \le
\frac{1 + (1 - \alpha)T + \alpha S}{(1 + T)(1 + S)}$$
$$\ln(1 + T) + \ln(1 + S) \le$$
$$\ln(1 + (1 - \alpha)S + \alpha T) + \ln(1 + (1 - \alpha)T + \alpha S)$$

Since $\ln(1 + x)$ is convex:
$$
\ln(1 + (1 - \alpha)S + \alpha T) \ge$$
$$(1 - \alpha)\ln(1 + S) + \alpha \ln(1 + T)$$
$$\ln(1 + (1 - \alpha)T + \alpha S) \ge (1 - \alpha)\ln(1 + T) + \alpha \ln(1 + S)$$

\end{document}